%% file: main.tex
\documentclass[11pt]{article}
\usepackage{float}
\usepackage[preprint]{acl}
\usepackage{placeins}

\usepackage{changepage}
\usepackage{rotating}
\usepackage{fancybox}
\usepackage{times}
\usepackage{printlen}
\usepackage{lipsum}
\usepackage{tcolorbox}
\usepackage{siunitx}
\usepackage{afterpage}
\usepackage{longtable}
\usepackage{adjustbox}
\newdimen\myfullpagetextwidth
\usepackage{latexsym}
\usepackage{booktabs, multirow, makecell}
\definecolor{mypurple}{HTML}{C05099}
\usepackage[T1]{fontenc}
\usepackage[english]{babel} 

\usepackage[utf8]{inputenc}
\usepackage{microtype}

\usepackage{inconsolata}

\usepackage{graphicx}
\usepackage{booktabs} % For better table formatting
\usepackage{multirow} % For multi-row cells in tables
\usepackage{amsmath}  % For math equations if needed

\usepackage{mdframed} % Use mdframed instead of tcolorbox

\definecolor{mypurple}{HTML}{C05099} % Ensure this is defined

\title{Mark My Words: A Robust Multilingual Model for Punctuation\\ in Text and Speech Transcripts}

\author{
  Sidharth Pulipaka$^{1,4}$\hspace{0.2cm}
  Sparsh Jain$^1$\hspace{0.2cm}
  Ashwin Sankar$^{1,2}$\hspace{0.2cm}
  \textbf{Raj Dabre}$^{1,2,3}$\thanks{Corresponding Author: \href{mailto:raj.dabre@cse.iitm.ac.in}{raj.dabre@cse.iitm.ac.in}}
  \\
  $^{1}$Nilekani Centre at AI4Bharat \quad
  $^{2}$Indian Institute of Technology, Madras \\
  $^{3}$Indian Institute of Technology, Bombay \quad
  $^{4}$Mahindra University, Hyderabad 
  \\
  \\
  \raisebox{-0.15em}{\includegraphics[height=0.9em]{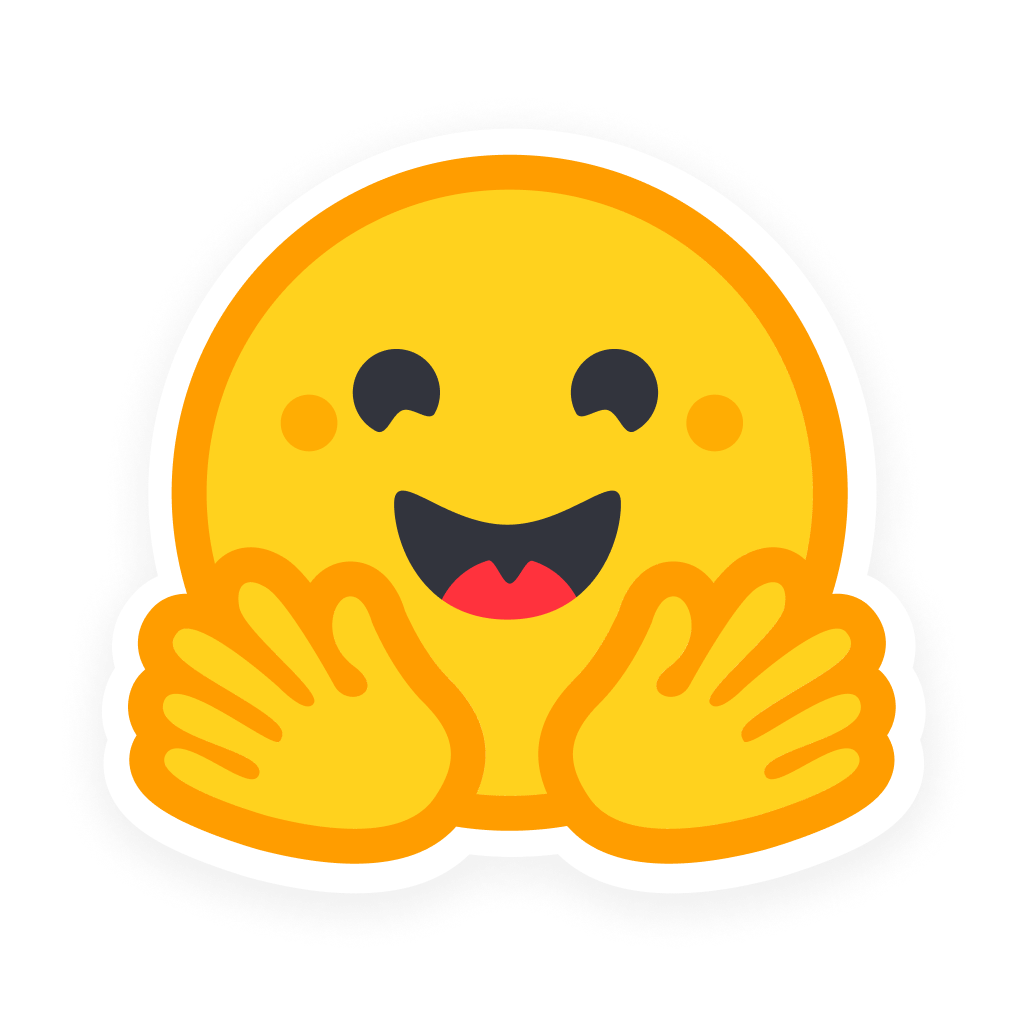}}~\small{\href{https://huggingface.co/ai4bharat/Cadence}{ai4bharat/Cadence}}
}

\begin{document}
\maketitle 

    % \begin{center} % Center the affiliation text
    %   \textsuperscript{1}Nilekani Centre at AI4Bharat
    %   \textsuperscript{2}Indian Institute of Technology, Bombay
    %   \textsuperscript{3}Indian Institute of Technology, Madras
    %   \textsuperscript{4}Google
    % \end{center}

\begin{abstract}
Punctuation plays a vital role in structuring meaning, yet current models often struggle to restore it accurately in transcripts of spontaneous speech, especially in the presence of disfluencies such as false starts and backtracking. These limitations hinder the performance of downstream tasks like translation, text-to-speech, summarization, etc. where sentence boundaries are critical for preserving quality. In this work, we introduce Cadence, a generalist punctuation restoration model adapted from a pretrained large language model. Cadence is designed to handle both clean written text and highly spontaneous spoken transcripts. It surpasses the previous state-of-the-art in performance while expanding support from 14 to all 22 Indian languages and English. We conduct a comprehensive analysis of model behavior across punctuation types and language families, identifying persistent challenges under domain shift and with rare punctuation marks. Our findings demonstrate the efficacy of utilizing pretrained language models for multilingual punctuation restoration and highlight Cadence’s practical value for low-resource NLP pipelines at scale.
\end{abstract}

\section{Introduction}

Punctuation plays a vital role in written language, offering syntactic structure, semantic clarity, and pragmatic cues such as pauses, emphasis, and sentence boundaries. However, text generated by Automatic Speech Recognition (ASR) systems or large-scale web crawls often lacks punctuation~\citep{10888018}. This absence significantly impairs both human readability and the effectiveness of downstream NLP tasks like Machine Translation (MT), Text Summarization, and Sentiment Analysis, posing a widespread challenge in processing raw textual data.

\begin{figure*}[t]
\centering
   \includegraphics[width=0.9\textwidth]{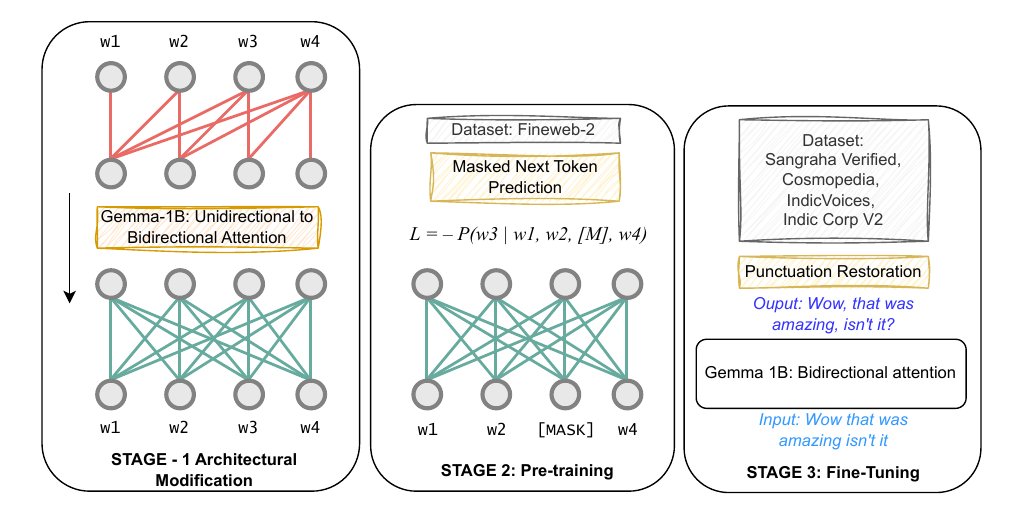}
    \caption{Overview of our training methodology. \textit{Stage-1:} Modify causal attention to bidirectional attention. \textit{Stage-2:} Pre-train with Masked Next Token Prediction Objective. \textit{Stage-3:} Train for punctuation restoration, as a token-level classification task. Figure inspired from \citealp{llm2vec}.}
    \label{fig:punctuation_figure}
\end{figure*}

While punctuation restoration has progressed for high-resource languages like English~\citep{devlin2019bertpretrainingdeepbidirectional}, Indic languages face substantial hurdles. These include scarcity of annotated corpora, especially for low-resource languages, and linguistic complexity with diverse scripts, grammars, and unique marks like the Devanagari danda. Prior efforts were often limited in language or punctuation scope, or used non-scalable, language-specific models, hindering cross-lingual generalization, particularly for under-represented languages~\citep{gupta2022indicpunctautomaticpunctuationrestoration}.

To bridge this gap for Indic languages, we first construct a large, diverse multilingual punctuation corpus covering both written and ASR-transcribed text. This corpus is aggregated from multiple sources, including Sangraha-verified~\cite{Khan_2024}, IndicVoices~\cite{javed2024indicvoicesbuildinginclusivemultilingual}, translated Cosmopedia~\cite{benallal2024cosmopedia}, and IndicCorp-v2~\cite{doddapaneni-etal-2023-towards}, and is curated to better balance linguistic representation, especially for low-resource languages, during training. Secondly, we adapt the \textsc{Gemma3-1B-pretrain} model \citep{gemmateam2025gemma3technicalreport} for punctuation restoration by converting it into a bidirectional transformer using a Masked Next Token Prediction (MNTP) objective \citep{llm2vec}. This adaptation enables efficient non-autoregressive sequence tagging, and our model -- Cadence supports a fine-grained label space of 30 distinct punctuation classes, encompassing standard punctuation, Indic-specific symbols, and frequent punctuation combinations, significantly extending the expressive capacity of prior approaches. 
Cadence achieves new state-of-the-art (SOTA) performance across multiple Indic languages and domains, surpassing existing baselines. Furthermore, we release Cadence to support downstream applications such as machine translation, speech translation, and large-scale text processing, with a particular focus on empowering NLP pipelines for under-resourced Indic languages.

Cadence supports English and all 22 scheduled languages of India: Assamese, Bengali, Bodo, Dogri, Gujarati, Hindi, Kannada, Kashmiri, Konkani, Maithili, Malayalam, Manipuri, Marathi, Nepali, Odia, Punjabi, Sanskrit, Santali, Sindhi, Tamil, Telugu, and Urdu.

In summary, our key contributions are: \textbf{(i)} the creation of an extensive multilingual Indic punctuation corpus, carefully curated from diverse sources to address data scarcity and improve linguistic representation for low-resource languages and \textbf{(ii)} we present an adapted Gemma3-based model, transformed into an efficient bidirectional sequence tagger via an MNTP objective, capable of restoring a comprehensive set of 30 punctuation classes, including Indic-specific symbols and common combinations.

\section{Related Work}

\noindent\textbf{Punctuation Restoration in Machine Translation and Speech Translation:} Punctuation restoration (PR) is a crucial preprocessing step for machine translation (MT) and speech translation (ST). In MT, punctuation provides essential segmentation and syntactic cues vital for translation quality \citep{vandeghinste-etal-2018-comparison}; its absence degrades translations. The impact is greater in ST, where unpunctuated Automatic Speech Recognition (ASR) transcripts hinder segmentation crucial for real-time systems and data alignment \citep{javed2024indicvoicesbuildinginclusivemultilingual, sankar2025towards}.

\noindent\textbf{Punctuation Restoration for Indic Languages:} Indic languages present unique PR challenges due to linguistic diversity and specific punctuation conventions. Early efforts were often monolingual, limiting scalability and cross-lingual transfer~\cite{tripathy-samal-2022-punctuation, gupta2022indicpunctautomaticpunctuationrestoration}. \citet{gupta2022indicpunctautomaticpunctuationrestoration} introduced IndicPunct, a multilingual transformer model for 14 Indian languages. While effective on formal text, IndicPunct faced limitations with spontaneous speech transcripts and a restricted punctuation set. These shortcomings highlight the need for more robust, generalist models for Indic languages, especially for spontaneous speech.

\noindent\textbf{Resources and Models for Punctuation Restoration:} State-of-the-art PR systems often use BERT-style token classifiers~\citep{gupta2022indicpunctautomaticpunctuationrestoration, deepMLP}, with Large Language Models (LLMs) recently gaining traction~\citep{sankar2025towards}. Earlier models, trained mainly on clean written text, struggle with disfluent spontaneous speech, impairing real-world ASR and ST applications. A key bottleneck is the scarcity of large, high-quality, punctuation-annotated corpora reflecting speech characteristics. While large multilingual text corpora like those by~\citep{penedo2024fineweb-2, doddapaneni-etal-2023-towards} support training for diverse Indic languages, they mostly contain formal or web text. Resources such as IndicVoices~\citep{javed2024indicvoicesbuildinginclusivemultilingual}, though unpunctuated, reveal stylistic phenomena PR models must address. However, non-autoregressive, generalist PR models for Indic languages that support large label sets and are robust across written and spoken styles remain rare. Cadence addresses this gap with a scalable, multilingual, LLM-based approach for varied domains and languages.

\section{Methodology}
\label{sec:methodology}
Our approach to developing a robust multilingual punctuation restoration model hinges on two core pillars: a comprehensive data strategy designed to encompass linguistic diversity and varied text styles, and a multi-stage model training and adaptation process.

\subsection{Data Strategy for Multilingual Punctuation Restoration}
The foundation of our methodology lies in the careful curation and utilization of diverse textual data for both pre-training and fine-tuning phases.

\noindent\textbf{Pre-training Data Corpus:}
For the initial continual pre-training phase, we leverage large-scale, high-quality multilingual web corpora. These resources are selected for their broad linguistic coverage, providing the model with exposure to a wide array of languages and writing styles. This extensive, general-domain data helps in building foundational representations that are adaptable to various downstream tasks and linguistic contexts.

\noindent\textbf{Fine-tuning Data Amalgamation:}
To prepare the model specifically for punctuation restoration, we construct a substantial and heterogeneous fine-tuning dataset. This is achieved through a significant data aggregation effort, where we consolidate textual data from a variety of distinct, existing corpora. These source datasets often vary in their original scale, domain focus (e.g., news, literature, web text, extempore text), and stylistic properties. The resulting amalgamated corpus intentionally includes both cleanly formatted, formal written text and less structured transcripts derived from spoken language. The inclusion of spoken language data, which may feature disfluencies, repetitions, and false starts, is crucial for training a model that is robust and performs well across a wide spectrum of real-world text input.

The primary goal of this data strategy is to expose the model to a rich tapestry of linguistic phenomena and punctuation usage patterns across numerous languages and text types, thereby fostering its ability to generalize effectively.

\subsection{Model Training and Adaptation}
The model undergoes a multi-stage training process, starting from a pre-trained foundation, followed by continual pre-training for domain and multilingual adaptation, and culminating in task-specific fine-tuning.

\subsubsection{Model Architecture Adaptation}
We begin with a foundation pre-trained transformer-based language model. Standard autoregressive language models are typically designed for unidirectional text generation, processing context only from preceding tokens. However, for sequence tagging tasks like punctuation restoration, where understanding the surrounding context is crucial, bidirectional information flow (accessing both preceding and succeeding tokens) is highly beneficial. Therefore, we adapt the model's attention mechanism to be fully bidirectional. This modification allows each token to attend to all other tokens in the input sequence, enabling a richer contextual understanding necessary for accurate punctuation prediction during subsequent training stages.

\subsubsection{Continual Pre-training for Enhanced Representation}
To further adapt the bidirectionally-modified model for the nuances of the diverse linguistic landscape it will encounter and to better prepare it for the sequence tagging nature of the punctuation restoration task, we perform a dedicated phase of continual pre-training.

\noindent\textbf{Masked Next Token Prediction Objective:}
Instead of conventional masked language modeling (MLM) or causal language modeling, we employ a \textit{Masked Next Token Prediction} (MNTP) objective \citep{llm2vec}. In this modified setup, a random subset of tokens in an input sequence is masked. The model's predictive task is then specifically focused: for an unmasked token at position $i$, if its subsequent token at position $i+1$ is masked, the model is trained to predict this masked token $i+1$. This prediction is performed using the contextual representation of the token at position $i$.

Crucially, the model employs bi-directional attention. This means the representation of token $i$ (which serves as the basis for predicting token $i+1$) is itself informed by the entire unmasked sequence, including tokens both preceding and succeeding token $i$. Despite access to this broader context, the objective's design hones the model's ability to leverage an unmasked token to specifically predict its immediate (masked) successor. This encourages the learning of strong local dependencies and makes the model adept at predicting upcoming elements based on their immediate preceding context - a skill particularly relevant for tasks like punctuation prediction, where marks typically follow a word or phrase.

\noindent\textbf{Curriculum Learning for Multilingual Adaptation:}
Given the significant variation in data availability (ranging from high-resource to low-resource languages) and the diverse linguistic characteristics across the target languages, we adopt a curriculum learning strategy during continual pre-training. This staged approach gradually introduces linguistic complexity to the model:
\begin{enumerate}
    \item \textbf{Foundation Phase:} Training initially focuses on a high-resource language (a language with abundant available training data) to establish robust foundational representations.
    \item \textbf{Expansion Phase 1 (Mid-to-High Resource):} The model is then exposed to a group of mid-to-high-resource languages. These are languages for which substantial amounts of training data are available, though typically less than the initial high-resource language. This phase allows the model to begin generalizing across related linguistic structures and benefit from these larger datasets.
    \item \textbf{Expansion Phase 2 (Low Resource):} Subsequently, lower-resource languages are introduced. These are languages characterized by comparatively limited availability of training data. This step encourages knowledge transfer from the more data-rich languages learned in previous phases, which is critical for achieving good performance on languages with scarce data.
    \item \textbf{Consolidation Phase:} Finally, the model is trained on a mixture of data from all supported languages. This phase aims to consolidate learning across the entire linguistic spectrum and mitigate potential catastrophic forgetting of earlier-learned languages or features.
\end{enumerate}
This progressive exposure helps the model to incrementally adapt to increasing linguistic diversity while maintaining training stability and fostering cross-lingual transfer.

\subsubsection{Task-Specific Fine-tuning for Punctuation Restoration}
Following the continual pre-training phase, the adapted language model is fine-tuned specifically for the punctuation restoration task using the amalgamated dataset described earlier.

\noindent\textbf{Task Formulation:} We frame punctuation restoration as a token-level sequence classification problem. For each token in an input unpunctuated sequence, the model's objective is to predict the punctuation mark that should follow it. If no punctuation is appropriate after a token, it predicts a special ``O" (Outside) label, indicating the absence of punctuation.

\noindent\textbf{Model Head:} To adapt the pre-trained model for this classification task, its original language modeling head (typically an unembedding matrix used for next token prediction) is replaced with a new task-specific head. This consists of a linear classification layer that takes the final hidden state representation of each input token and outputs logits over the set of possible punctuation classes.

\noindent\textbf{Punctuation Label Space:} The model is trained to predict a comprehensive set of punctuation marks. This set includes standard punctuation common across many languages, as well as punctuation specific to particular scripts or linguistic traditions found within the target languages. It also covers frequently occurring combinations of punctuation marks to capture more complex typographic conventions.

\noindent\textbf{Sampling Strategy for Data Imbalance:} Recognizing the varying amounts of data available for different languages within our fine-tuning corpus, we employ a weighted sampling strategy during this final training stage. This technique ensures that lower-resource languages, which inherently have fewer training examples, are adequately represented in training batches. By oversampling data from these languages, we aim to provide a more equitable training signal, promoting more balanced learning and fostering robust performance across all targeted languages, irrespective of their individual data abundance.

\section{Experimental Setup}
This section outlines the datasets, training procedures, and evaluation metrics used to develop and assess Cadence.

\subsection{Datasets} 
The development of Cadence relies on carefully curated datasets for both its continual pre-training and task-specific fine-tuning phases, ensuring broad linguistic coverage and exposure to diverse text styles.

\subsubsection{Pretraining Dataset}
We source pretraining data from the Indic subset of FineWeb-2~\citep{penedo2024fineweb-2}. This high-quality, multilingual web corpus provides broad coverage across the Indian linguistic landscape, a result of its web-scale collection and rigorous filtering.

\label{app:data_stats}
% First Table - Language Data
\begin{table}[htbp]
    \footnotesize
    \centering
    \begin{tabular}{@{} lr @{}}
        \toprule
        \textbf{Language} & \ \ \textbf{Number of} \\ & \textbf{Samples} \\
        \midrule
        Assamese   & 106k \\
        Bengali    & 126k \\
        Bodo       & 36k  \\
        Dogri      & 3k   \\
        English    & 89k  \\
        Gujarati   & 102k \\
        Hindi      & 123k \\
        Kannada    & 110k \\
        Kashmiri   & 35k  \\
        Konkani    & 42k  \\
        Maithili   & 46k  \\
        Malayalam  & 125k \\
        Marathi    & 126k \\
        Nepali     & 127k \\
        Odia       & 106k \\
        Punjabi    & 104k \\
        Sanskrit   & 75k  \\
        Santali    & 66k  \\
        Sindhi     & 58k  \\
        Tamil      & 122k \\
        Telugu     & 115k \\
        Urdu       & 123k \\
        \midrule
        \textbf{Total}    & \textbf{1,965k} \\
        \bottomrule
    \end{tabular}
    \caption{Statistics of our training corpus, showing the number of entries available for each supported language, represented in thousands.}
    \label{tab:language_data}
\end{table}

\begin{table}[t]
\centering
   % Specify a height, e.g., a fraction of the text height
   % Try values like 0.3\textheight, 0.4\textheight, etc.
   \includegraphics[height=0.5\textheight] % Example: 30% of text height
   {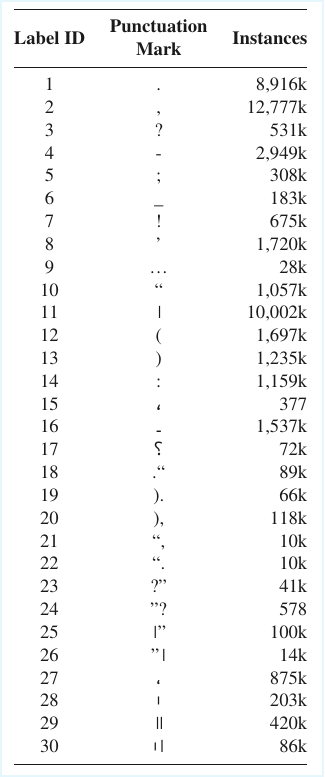}

    \caption{Breakdown of supported punctuation marks, their internal Label IDs, and the number of instances for each in our training corpus, represented in thousands. For language wise breakdown, refer to Appendix \ref{subsec:lang_wise_breakdown}}
    \label{fig:Punctuation_mark_table}
\end{table}

\subsubsection{Fine-tuning Datasets for Punctuation Restoration}
\label{sec:fine_tuning_datasets}
To effectively fine-tune our punctuation restoration model, we compiled a substantial multilingual dataset by aggregating data from four diverse sources. This approach ensures broad linguistic coverage and exposure to various punctuation patterns. The constituent datasets are detailed below:

\noindent\textbf{Sangraha-Verified (S):} 
This dataset, based on the work of \citet{Khan_2024}, consists of web-scraped text that has undergone a verification process. For punctuation restoration, this implies a higher quality of accurately punctuated sentences, providing reliable examples for the model to learn correct punctuation usage.

\noindent\textbf{IndicVoices-ST (IV):}
IndicVoices-ST \citep{sankar2025towards} is primarily a speech-to-text corpus for Indic languages. This dataset is invaluable as it contains transcribed speech where punctuation was added using \textsc{Llama-3-405B-Instruct} (prompt in \ref{subsec:punctuation_prompt}), offering training material that reflects spoken language patterns in their punctuated written forms.

\noindent\textbf{Translated Cosmopedia (C):}
We expanded our data by translating Cosmopedia \citep{benallal2024cosmopedia}, a large-scale general knowledge dataset, into various Indic languages using the \textsc{Llama-3-405B-Instruct} model \citep{grattafiori2024llama3herdmodels}. This provides a significant volume of diverse, structured text, where the translated punctuation, while potentially imperfect, broadens the model's exposure to different contexts and syntactic structures across languages.

\noindent\textbf{IndicCorp-v2 (IC):}
IndicCorp-v2 \citep{doddapaneni-etal-2023-towards} is a comprehensive corpus of text for Indic languages. Its large scale and wide domain coverage contribute a vast number of natural language examples, helping the model learn common and nuanced punctuation conventions prevalent in these languages.

By combining these sources, we created a robust training corpus. Table \ref{tab:language_data} provides a detailed breakdown of this aggregated dataset, showing the approximate number of training instances per language and Table \ref{fig:Punctuation_mark_table} shows the counts for each of the 30 supported punctuation labels in the overall training set.

\begin{table*}[!h] % Use table* for a wide table spanning two columns in a two-column document
\centering
% \scriptsize
\fontsize{8pt}{10pt}\selectfont
\setlength{\tabcolsep}{3pt} % Reduce column separation (default is usually 6pt)
\begin{tabular}{l | r r | r r r r r | r r r r | r c} % Removed vertical bars, specified 14 columns
\toprule
\multirow{2}{*}{\textbf{Language}} & \multicolumn{2}{c}{\begin{tabular}[l]{@{}c@{}}\textbf{Number of} \\ \textbf{Samples}\end{tabular}} & \multicolumn{5}{c}{\textbf{Cadence (Ours)}} & \multicolumn{4}{c}{\textbf{IndicPunct}} & \multicolumn{2}{c}{\textbf{DMP}} \\
\cmidrule(lr){2-3} \cmidrule(lr){4-8} \cmidrule(lr){9-12} \cmidrule(lr){13-14} % Use cmidrule for partial rules
 & \textbf{Formal} & \textbf{Extempore} & \textbf{S} & \textbf{IC} & \textbf{C} & \textbf{BPCC} & \textbf{IV} & \textbf{S} & \textbf{IC} & \textbf{C} & \textbf{IV} & \textbf{IC} & \textbf{BPCC} \\ % Note: Adjusted header for DeepMultiPunct to match the third table
\midrule % Main horizontal line below the header
Assamese   & 1,426 & 1,275 & \textbf{0.71} & \textbf{0.76} & \textbf{0.81} & \textbf{--}     & \textbf{0.60} & 0.30  & 0.48 & 0.24 & 0.48 & x   & x   \\
Bengali    & 1,499 & 1,447 & \textbf{0.54} & \textbf{0.72} & \textbf{0.84} & \textbf{--}     & \textbf{0.60} & 0.30  & 0.54 & 0.20  & 0.42 & x   & x   \\
Bodo       & 1,057 &  860 & \textbf{--}     & \textbf{0.75} & \textbf{--}     & \textbf{0.42} & \textbf{0.29} & x    & x    & x    & x    & x   & x   \\
Dogri      &  919 &  995 & \textbf{--}     & \textbf{0.52} & \textbf{--}     & \textbf{0.42} & \textbf{0.30} & x    & x    & x    & x    & x   & x   \\
Gujarati   & 1,479 & 1,063 & \textbf{0.58} & \textbf{0.64} & \textbf{0.80} & \textbf{--}     & \textbf{0.54} & 0.19 & 0.33 & 0.19 & 0.34 & x   & x   \\
Hindi      & 1,669 & 1,273 & \textbf{0.61} & \textbf{0.76} & \textbf{0.84} & \textbf{--}     & \textbf{0.65} & 0.34 & 0.5  & 0.23 & 0.46 & x   & x   \\
Kannada    & 1,473 & 1,165 & \textbf{0.60} & \textbf{0.79} & \textbf{0.77} & \textbf{--}     & \textbf{0.61} & 0.25 & 0.45 & 0.19 & 0.41 & x   & x   \\
Konkani    &  994 &  993 & \textbf{0.78} & \textbf{0.61} & \textbf{--}     & \textbf{0.32} & \textbf{0.37} & x    & x    & x    & x    & x   & x   \\
Maithili   &  984 &  998 & \textbf{0.64} & \textbf{0.73} & \textbf{--}     & \textbf{0.50} & \textbf{0.40} & x    & x    & x    & x    & x   & x   \\
Malayalam  & 1,532 & 1,270 & \textbf{0.67} & \textbf{0.74} & \textbf{0.77} & \textbf{--}     & \textbf{0.69} & 0.29 & 0.43 & 0.22 & 0.41 & x   & x   \\
Marathi    & 1,786 & 1,216 & \textbf{0.73} & \textbf{0.74} & \textbf{0.82} & \textbf{--}     & \textbf{0.56} & 0.21 & 0.35 & 0.22 & 0.49 & x   & x   \\
Nepali     & 1,111 &  954 & \textbf{0.69} & \textbf{0.78} & \textbf{--}     & \textbf{--}     & \textbf{0.51} & x    & x    & x    & x    & x   & x   \\
Odia       & 1,341 & 1,723 & \textbf{0.72} & \textbf{0.77} & \textbf{0.68} & \textbf{--}     & \textbf{0.72} & 0.28 & 0.44 & 0.20  & 0.38 & x   & x   \\
Punjabi & 1,424 & 1,322 & \textbf{0.70} & \textbf{0.71} & \textbf{0.69} & \textbf{--} & \textbf{0.48} & 0.36 & 0.43 & 0.32 & 0.51 & x & x \\
Sanskrit   & 1,118 &  983 & \textbf{0.23} & \textbf{0.51} & \textbf{--}     & \textbf{0.43} & \textbf{0.35} & x    & x    & x    & x    & x   & x   \\
Santali    &  443 &  575 & \textbf{--}     & \textbf{0.79} & \textbf{--}     & \textbf{--}     & \textbf{0.37} & x    & x    & x    & x    & x   & x   \\
Sindhi     & 1,277 &  947 & \textbf{0.52} & \textbf{0.50} & \textbf{--}     & \textbf{0.33} & \textbf{0.37} & x    & x    & x    & x    & x   & x   \\
Tamil      & 1,447 & 1,369 & \textbf{0.65} & \textbf{0.72} & \textbf{0.78} & \textbf{--}     & \textbf{0.59} & 0.25 & 0.58 & 0.20  & 0.44 & x   & x   \\
Telugu     & 1,451 & 1,308 & \textbf{0.76} & \textbf{0.74} & \textbf{0.80} & \textbf{--}     & \textbf{0.54} & 0.23 & 0.4  & 0.19 & 0.32 & x   & x   \\
Urdu       & 1,562 & 1,252 & \textbf{0.65} & \textbf{0.72} & \textbf{0.64} & \textbf{--}     & \textbf{0.74} & x    & x    & x    & x    & x   & x   \\
Kashmiri   & 1,259 &  981 & \textbf{--}     & \textbf{0.66} & \textbf{--}     & \textbf{0.52} & \textbf{0.33} & x    & x    & x    & x    & x   & x   \\
Manipuri   & 1,074 & --   & \textbf{--}     & \textbf{--}     & \textbf{--}     & \textbf{0.44} & \textbf{--}     & x    & x    & x    & x    & x   & x   \\
English    & 1,035 & --   & \textbf{--}     & 0.54 & \textbf{--}     & \textbf{0.63} & \textbf{--}     & x    & x    & x    & x    & 0.54 & 0.50  \\
\midrule
\textbf{Overall} & 29,360 & 23,969 & \textbf{0.68} & \textbf{0.76} & \textbf{0.78} & \textbf{0.45} & \textbf{0.60} & 0.31 & 0.54 & 0.26 & 0.54 & 0.54 & 0.50\\
\bottomrule
\end{tabular}
\caption{Comparison of Punctuation Restoration Model Performance Across Languages and Metrics. An x indicates that the model does not support the given language. A -- indicates that results are unavailable due to insufficient high-quality data samples. All scores are reported on Focus Labels for consistency and comparability. DMP stands for DeepMultilingualPunctuation.}
\label{tab:combined_comparison}
\end{table*}

\subsection{Training Details}
In this section we elucidate the training details including model architecture, pretraining and finetuning details and evaluation setup.
\subsubsection{Model Architecture}
We adopt \textsc{Gemma3-1B-pretrain}~\citep{gemmateam2025gemma3technicalreport} as our base model. Although Gemma was originally designed as a causal decoder for text generation, punctuation restoration benefits from access to bidirectional context. We modify the \textsc{Gemma-3-1B}'s attention mechanism to attend to both left and right contexts, thus making it bidirectional.

\subsubsection{Continual Pretraining}
To adapt the modified \textsc{Gemma-3-1B} model to the multilingual and low-resource nature of the task, we perform continual pretraining using a masked next token prediction \citep{llm2vec} objective and a curriculum over languages.

% \noindent\textbf{Masked Next Token Prediction:} 
% Rather than standard masked language modeling (MLM) or causal prediction, we adopt the \textit{Masked Next Token Prediction} (MNTP) objective~\citep{llm2vec}. Given a sequence, we randomly mask a subset of tokens. For each masked token at position $i+1$, the model is trained to predict the masked token at the position of its previous token $i$. This encourages learning short-range, forward-predictive dependencies, which are crucial for modeling punctuation, particularly after disfluent or spontaneous phrases in speech transcripts.

\noindent\textbf{Curriculum Learning Strategy:} 
Given the wide variation in data availability and linguistic structure across Indic languages, we employ a four-phase curriculum learning strategy:

\noindent\textit{Phase 1: English only} -- Initializes the model with a high-resource language to establish stable representations. masking ratio was set to 0.30.

\noindent\textit{Phase 2: High- and mid-resource Indic languages} -- Introduces 13 languages, which includes Hindi, Telugu, Tamil, Bengali, Malayalam, Marathi, Kannada, Gujarati, Assamese, Oriya, Punjabi, Sindhi, Urdu. 0.25 masking ratio was employed.

\noindent\textit{Phase 3: Low-resource Indic languages} -- Adds the languages: Bodo, Dogri, Konkani, Kashmiri, Maithili, Manipuri, Nepali, Sanskrit, Santali, encouraging generalization. Masking ratio was 0.15.

\noindent\textit{Phase 4: Mixed multilingual training} -- We train on all 23 languages (22 Indic languages + English) for the final 10\% of steps to consolidate knowledge and mitigate catastrophic forgetting. 0.25 masking ratio was used.

This staged progression allows the model to incrementally adapt to increasing linguistic diversity while maintaining stability across training phases.

\subsubsection{Finetuning}
After the pretraining stage, the model is fine-tuned for supervised punctuation restoration, framed as token-level sequence tagging. For each input token, it predicts a subsequent punctuation mark or an O (Outside) label if no punctuation follows. To achieve this, \textsc{Gemma-3-1B}'s original language modeling head is replaced with a linear classification layer that operates on final token hidden states, outputting logits over the 30 punctuation classes. This comprehensive label space (detailed in Table~\ref{tab:punctuation_data}) includes standard English, Indic-specific, Arabic-script (Urdu) marks, and frequent multi-character combinations, enabling fine-grained modeling of diverse writing styles. \newline We trained Cadence using the AdamW optimizer \citep{loshchilov2017decoupled} with a max learning rate of $2e-4$ with 10\% of the training steps as warmup followed by a cosine decay to $1e-6$. The effective batch size was 64 and the model was trained on 8$\times$H100 GPUs for 8 hours.

\subsection{Evaluation}
\noindent\textbf{Test Set:}
We held out test set instances from IndicCorp-v2, Sangraha-Verified, translated Cosmopedia dataset and IndicVoices which has spontaneous speech transcripts. Apart from these, we have also used the BPCC dataset from \citealp{gala2023indictrans}. Ensuring the quality of punctuation in ground truth test data is crucial for reliable evaluation. We employed Google's Gemini-2.5-Flash-preview-04-17 model as a judge to automatically assess the quality of punctuation in our held-out test set instances. Using a detailed rubric (prompt in Appendix \ref{subsec:llm_as_a_judge}) focusing on grammatical correctness, appropriateness, and necessity of punctuation marks, we assigned scores from 1 to 5. Only instances scoring 4.5 or higher were retained in the final test set, ensuring a high-quality benchmark. \\

\noindent\textbf{Evaluation Metric:}
We evaluate the model's performance using the Average Macro F1 score. This metric calculates the F1 score for each punctuation class independently and then averages them, giving equal weight to each class, which is suitable for imbalanced class distributions common in punctuation restoration. \\

\noindent\textbf{Baselines:}
We compare Cadence with the following baseline models:

\textbf{(i)} \textit{IndicPunct} \citep{gupta2022indicpunctautomaticpunctuationrestoration}: This series of models consists of language-specific fine-tuned versions of IndicBert \citep{kakwani2020indicnlpsuite}, designed for punctuation restoration. However, IndicPunct supports a limited set of languages and a restricted label space, specifically sentence-end marks, question marks, and commas.

\textbf{(ii)} \textit{Deepmultilingualpunctuation} \citep{deepMLP}: This model is trained to restore punctuation in European languages. We utilize it as a baseline for English. Similar to IndicPunct, Deepmultilingualpunctuation supports a limited label space, including full stops, question marks, commas, hyphens, and colons.

To ensure fair comparisons across these models with differing label scopes, we evaluate performance on a common, limited set of labels, which we term "Focus labels." \emph{The ``focus labels`` include the period (.), comma (,), colon (:), question mark (?), Devanagari danda, Urdu full stop, Urdu question mark, Arabic commas, and Santali mucaad.}

\section{Results}
\label{sec:results}

This section evaluates Cadence's performance. We first compare its efficacy on a defined set of ``focus labels" versus all supported punctuation labels, alongside a comparison with baseline models. We then analyze its performance across formal written text and spontaneous extempore transcripts. This is followed by an examination of the relationship between training data volume and performance, and finally, its generalization capabilities to unseen languages.

\subsection{Performance on Focus vs. All Labels and Baseline Comparison}
\label{ssec:focus_vs_all_and_baselines}

Cadence demonstrates strong performance on critical punctuation marks. When evaluated on ``focus labels", it achieves an overall score of \textbf{0.7924} on written text and \textbf{0.6249} on spontaneous speech transcripts (Table~\ref{tab:results_full_set}).

This achieves a substantial improvement over existing baselines. As presented in Table~\ref{tab:combined_comparison}, Cadence attains a score of \textbf{0.76} on IndicCorp-v2, markedly surpassing both IndicPunct (0.54) and DeepMultilingualPunctuation (0.54). Similarly, Cadence demonstrates superior performance on Sangraha-Verified (0.68 vs. 0.31), Translated Cosmopedia (0.78 vs. 0.26), and the IndicVoices dataset (0.60 vs. 0.54). On the BPCC dataset for English, Cadence achieves a score of 0.63, outperforming DeepMultilingualPunctuation (0.50), while additionally offering broad language coverage across numerous Indic languages that are unsupported by the existing baselines.

When considering the complete set of 30 supported punctuation labels (Table~\ref{tab:results_full_set}), Cadence achieves an overall score of \textbf{0.5931} on written text and \textbf{0.4508} on spontaneous speech transcripts. These lower scores, relative to the subset of focus labels, are expected given that the broader label set includes rarer punctuation marks, those exhibiting greater stylistic variation, and marks reflecting more nuanced syntactic or prosodic boundaries - all of which substantially increase the task’s complexity.

\subsection{Performance on Formal Written Text vs. Extempore Transcripts}
\label{ssec:formal_vs_extempore}

Cadence consistently performs better on formal written text than on extempore speech transcripts. For ``focus labels", the overall score is \textbf{0.7924} for written text versus \textbf{0.6249} for spontaneous speech transcripts (Table~\ref{tab:results_full_set}). A similar trend is observed for ``all labels", with scores of \textbf{0.5931} (written) and \textbf{0.4508} (extempore).

Furthermore, spontaneous utterances often present reduced syntactic regularity, featuring fragmented constructions and anacolutha, making the automatic identification of logical punctuation points ambiguous.  These difficulties are notably exacerbated by the comparatively limited availability of large, diverse, and accurately annotated training corpora for spontaneous speech transcripts, hindering the model's ability to learn robust patterns for these irregular linguistic phenomena.

% These characteristics, along with potential ASR errors and prosodic cues (like pauses) that may not be perfectly transcribed or aligned with standard punctuation, make accurate punctuation restoration more challenging compared to cleaner, well-structured written text.

\subsection{Impact of Training Data Volume on Performance}
\label{ssec:data_vs_performance}

An analysis of fine-tuning data volume (Table \ref{tab:language_data}) in relation to model performance on the written transcript test set with all punctuation labels (Table \ref{tab:results_full_set}) reveals a broad, though not strictly linear, correlation. Languages with larger fine-tuning datasets tend to achieve stronger F1 scores. For instance, Bengali (126k samples, 0.5087), Hindi (123k, 0.4966), and Telugu (115k, 0.5349) all demonstrate solid performance. Assamese (106k, 0.5324) and Marathi (126k, 0.4961) similarly perform well. 

In contrast, languages with much smaller fine-tuning corpora show lower performance: Dogri (3k samples, 0.2432) and Manipuri (4k, 0.2822) are notable examples. However, fine-tuning volume alone does not fully account for the performance spread. For example, Sanskrit (75k, 0.2966) and Sindhi (58k, 0.2966) perform poorly despite having moderate data, likely reflecting limited representation in the base \textsc{Gemma-3-1B} model’s original pretraining. In contrast, English (89k, 0.5125) and Punjabi (104k, 0.4849) achieve better outcomes, likely benefiting from richer pretraining exposure.

\begin{table}[t]
\centering
\fontsize{9pt}{12pt}\selectfont
\begin{tabular}{l|rr|rr}
\toprule
\textbf{Language} & \multicolumn{2}{c}{\textbf{Written}} & \multicolumn{2}{c}{\textbf{Extempore}} \\
\cmidrule(lr){2-3} \cmidrule(lr){4-5}
& \begin{tabular}[l]{@{}c@{}}\textbf{All} \\ \textbf{Labels}\end{tabular} & \textbf{\begin{tabular}[l]{@{}c@{}}\textbf{Focus} \\ \textbf{Labels}\end{tabular}} & \textbf{\begin{tabular}[l]{@{}c@{}}\textbf{All} \\ \textbf{Labels}\end{tabular}} & \textbf{\begin{tabular}[l]{@{}c@{}}\textbf{Focus} \\ \textbf{Labels}\end{tabular}} \\
\midrule
Assamese & 0.53 & 0.80 & 0.42 & 0.60 \\
Bengali & 0.50 & 0.70 & 0.38 & 0.60 \\
Bodo & 0.38 & 0.58 & 0.31 & 0.29 \\
Dogri & 0.24 & 0.42 & 0.27 & 0.30 \\
English & 0.38 & 0.59 & ---- & ---- \\
Gujarati & 0.50 & 0.67 & 0.43 & 0.54 \\
Hindi & 0.49 & 0.82 & 0.38 & 0.65 \\
Kannada & 0.45 & 0.65 & 0.40 & 0.61 \\
Kashmiri & 0.32 & 0.57 & 0.23 & 0.33 \\
Konkani & 0.36 & 0.57 & 0.18 & 0.37 \\
Maithili & 0.36 & 0.59 & 0.27 & 0.40 \\
Malayalam & 0.51 & 0.67 & 0.34 & 0.69 \\
Manipuri & 0.26 & 0.44 & ---- & ---- \\
Marathi & 0.49 & 0.82 & 0.43 & 0.56 \\
Nepali & 0.44 & 0.73 & 0.36 & 0.51 \\
Odia & 0.42 & 0.71 & 0.44 & 0.72 \\
Punjabi & 0.45 & 0.68 & 0.40 & 0.48 \\
Sanskrit & 0.21 & 0.35 & 0.20 & 0.35 \\
Santali & 0.58 & 0.79 & 0.20 & 0.37 \\
Sindhi & 0.29 & 0.45 & 0.24 & 0.37 \\
Tamil & 0.50 & 0.76 & 0.30 & 0.59 \\
Telugu & 0.53 & 0.79 & 0.35 & 0.54 \\
Urdu & 0.46 & 0.68 & 0.73 & 0.76 \\
\midrule
\textbf{Total} & \textbf{0.59} & \textbf{0.79} & \textbf{0.45} & \textbf{0.63}\\
\bottomrule
\end{tabular}
\caption{Cadence: Per-language Macro F1 Scores on Written and spontaneous speech transcripts test sets, evaluated on all 30 punctuation labels.}
\label{tab:results_full_set}
\end{table}

Linguistic and script complexity also play roles. Sanskrit, for instance, may suffer due to morphophonological processes such as sandhi, which complicate punctuation modeling. Interestingly, Santali (66k, 0.5864) significantly outperforms several better-resourced languages. This may be due to favorable transfer effects during multilingual continual pretraining or distinct punctuation conventions that align better with the model’s learned representations.

\subsection{Generalization to Unseen and Low-Resource Languages}
\label{ssec:generalization_unseen_low_resource} % Kept original label as content is similar

We investigated Cadence's ability to generalize to languages and conditions not extensively covered during fine-tuning.

\paragraph{Zero-Shot Generalization on Bhojpuri:}
Cadence was evaluated on Bhojpuri text from FineWeb-2 \cite{penedo2024fineweb-2}. Bhojpuri was absent from Cadence's continual pre-training and fine-tuning, though the base \textsc{Gemma-3-1B-pretrain} model \citep{gemmateam2025gemma3technicalreport} had prior exposure. In this zero-shot setting, Cadence achieved a Macro F1 score of \textbf{0.4615} on ``focus labels", indicating a promising capability for unseen language adaptation. 

\paragraph{Low-Resource Performance on Manipuri:}
Manipuri served as a low-resource test case, with minimal training samples. Compounding this, the \textsc{Gemma-3-1B-pretrain} tokenizer's lack of native Meitei Mayek script support necessitated using Bengali-script transliterated Manipuri for training and evaluation. Despite these constraints, Cadence scored 0.44 (on ``focus labels", BPCC dataset, Table~\ref{tab:combined_comparison}) and 0.2684 (all labels, written text, Table~\ref{tab:results_full_set}). These findings underscore Cadence's potential in low-data scenarios, even with script-related challenges.

\section{Conclusion}

We presented Cadence, a novel multilingual punctuation restoration model for English and 22 scheduled Indian languages. By adapting the \textsc{Gemma3-1B-pretrain} model with bidirectional attention and utilizing a curriculum-based continual pre-training strategy with MNTP on Indic web data, we successfully created a robust foundation model. Fine-tuning this model on a diverse aggregation of datasets with weighted sampling yielded a single model capable of handling 23 languages and 30 punctuation types, including Indic-specific marks. Our model significantly outperforms existing monolingual baselines across various languages, demonstrating the power of multilingual learning and our tailored pre-training approach.This achievement highlights the potential of unified multilingual models to address linguistic disparities in the realm.

\section{Limitations}
Despite Cadence's strong performance, several limitations should be acknowledged. Firstly, the quality and representativeness of the training data, particularly for low-resource languages and spontaneous speech, remain a challenge. This directly impacts performance, with the model exhibiting weaker results on languages for which less training data was available. Secondly, while our 30-label set is comprehensive, it may not capture extremely rare or highly nuanced stylistic punctuation. Finally, performance on spontaneous speech, though improved, still lags behind that on formal text, highlighting the persistent difficulty of modeling the irregularities and disfluencies inherent in spoken language. 

\section{Future Work}
\label{sec:future_work}

Looking ahead, we plan to conduct extensive evaluations to quantify the impact of Cadence's punctuation restoration on various downstream NLP tasks. Specifically, we aim to investigate its effects on the performance of Machine Translation (MT), Text-to-Speech (TTS) synthesis, and the accuracy of Automatic Speech Recognition (ASR) systems, particularly for Indic languages. This will help to further establish the practical benefits of robust punctuation in real-world NLP pipelines.

% Custom bibliography entries only
\bibliography{custom,references} % You need to create a custom.bib file

\appendix
\newpage
\input{sections/appendix}
\end{document}

%% file: sections/appendix.tex
\newdimen\myfullpagetextwidth

\onecolumn
\section*{Appendix} \label{sec:appendix}

\FloatBarrier

\renewcommand{\topfraction}{0.9}
\renewcommand{\bottomfraction}{0.8}
\setcounter{topnumber}{2}
\setcounter{bottomnumber}{2}
\setcounter{totalnumber}{4}
\renewcommand{\dbltopfraction}{0.9}
\renewcommand{\textfraction}{0.07}
\renewcommand{\floatpagefraction}{0.7}

\section{Language wise label breakdown}
\label{subsec:lang_wise_breakdown}
The tables show language wise breakdown of label distribution.

\begin{table}[H]
    \centering
   \includegraphics[scale=0.8]
   {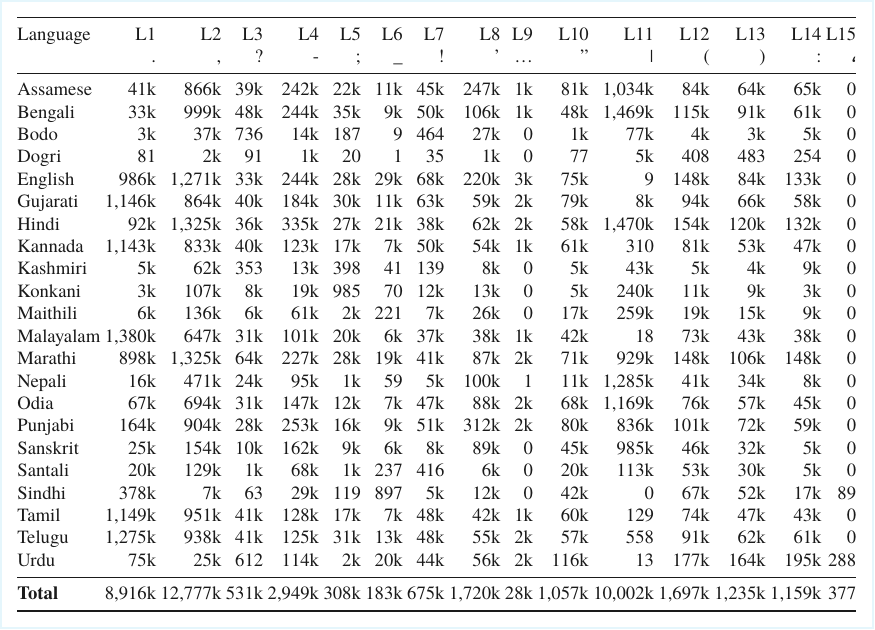}
   
    \caption{Label Distribution per Language (Part 1: L1-L15). Counts $\ge$ 1000 are shown in thousands (k). Top header row is Label ID, second header row is the corresponding punctuation mark.}
    \label{tab:punctuation_lang_count1}
\end{table}

\FloatBarrier

\begin{table}[H]
    \centering
   \includegraphics[width=0.73\textwidth]
   {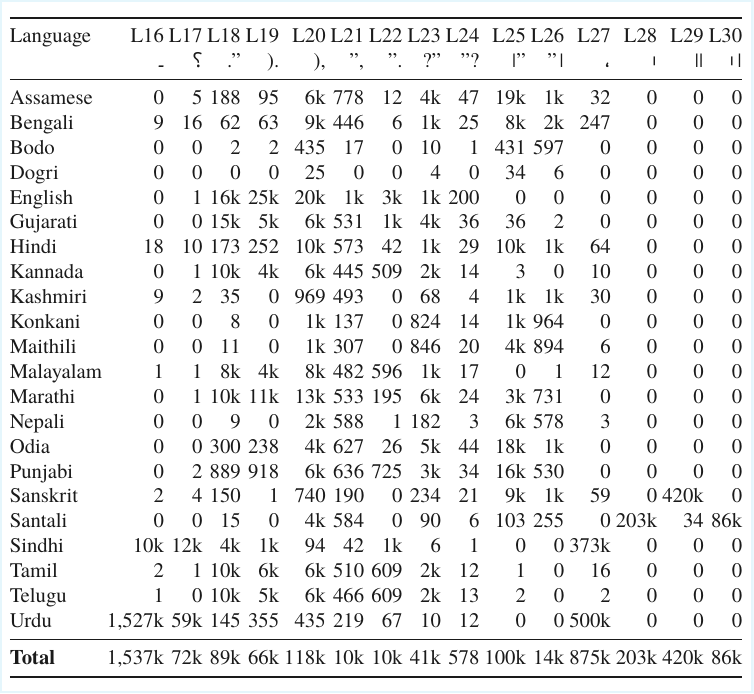}
   
    \caption{Label Distribution per Language (Part 2: L16-L30). Counts $\ge$ 1000 are shown in thousands (k). Top header row is Label ID, second header row is the corresponding punctuation mark.}
    \label{tab:punctuation_lang_count2}
\end{table}

\twocolumn
\clearpage
\section{Prompt used for Punctuation} 
\label{subsec:punctuation_prompt}
% The prompt template illustrated in Figure is designed to generate appropriate punctuation marks and facilitate the construction of data for punctuation restoration tasks.

\begin{figure}[htbp]
\centering
   \includegraphics[width=\columnwidth]
   {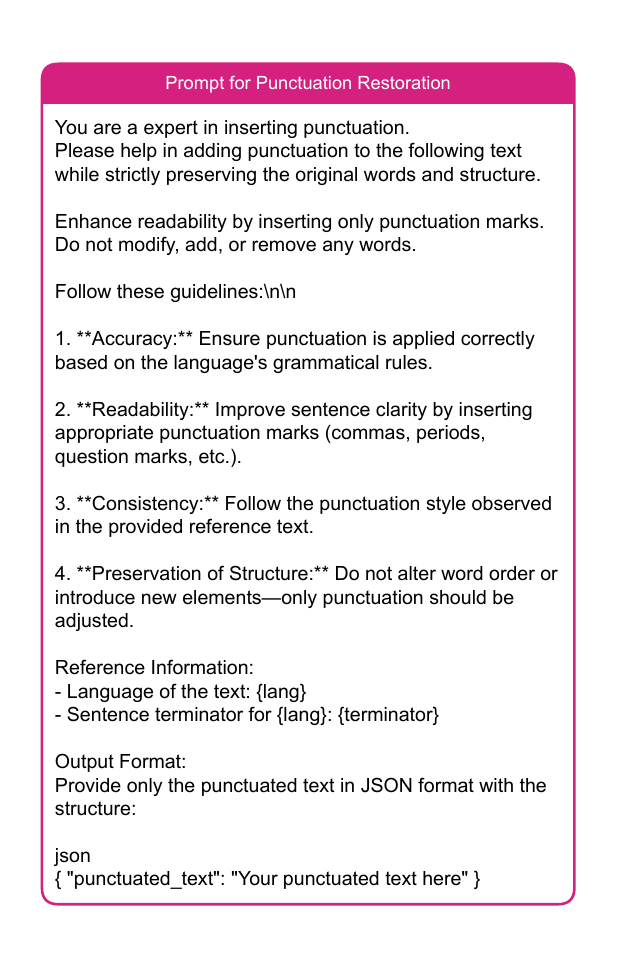}
   \vspace{-20pt}
    \caption{Prompt For Punctuation Restoration}
    \label{fig:Prompt_Punctuation}
\end{figure}

The prompt template shown in Figure~\ref{fig:Prompt_Punctuation} is engineered to guide Large Language Models (LLMs) in the task of punctuation restoration for Indian language text. It begins by defining the LLM's role as a punctuation expert and sets a primary objective: to enhance text readability by inserting punctuation marks while strictly preserving the original wording and sentence structure.

The prompt enumerates four critical guidelines for the LLM:
\begin{enumerate}
    \item \textbf{Accuracy:} Punctuation must conform to the grammatical rules of the specified input language ({lang}).
    \item \textbf{Readability:} Sentence clarity should be improved using appropriate punctuation (e.g., commas, periods, question marks).
    \item \textbf{Consistency:} The punctuation style should align with any provided reference text.
    \item \textbf{Preservation of Structure:} Word order and sentence construction must remain unaltered; only punctuation is to be adjusted.
\end{enumerate}

To accommodate linguistic diversity, particularly the varied sentence terminators across Indian languages (e.g., period vs. danda), the prompt requires the input language ({lang}) and its corresponding sentence terminator ({terminator}) as explicit parameters. Finally, it mandates a structured JSON output with the key "punctuated\_text", ensuring the punctuated text is returned in a consistent, machine-readable format. This design facilitates systematic generation of punctuated data suitable for training and evaluating punctuation restoration models.

\newpage

\section{Prompt used for LLM as a Judge} 
The datasets we have used for training contain web-scraped text (Sangraha-verified, IndicCorpV2) and also synthetically punctuated text (IndicVoices). As a result punctuations may not always be correct. Ensuring a high quality test set becomes important to accurately assess our model and compare with existing models. 
We have employed Gemini-2.5-Flash-preview-04-17 as a judge to validate our test set. We present the prompt used in Fig.\ref{fig:llm_as_a_judge} below.
\label{subsec:llm_as_a_judge}

\begin{figure*}
    \includegraphics[width=1.0\linewidth]{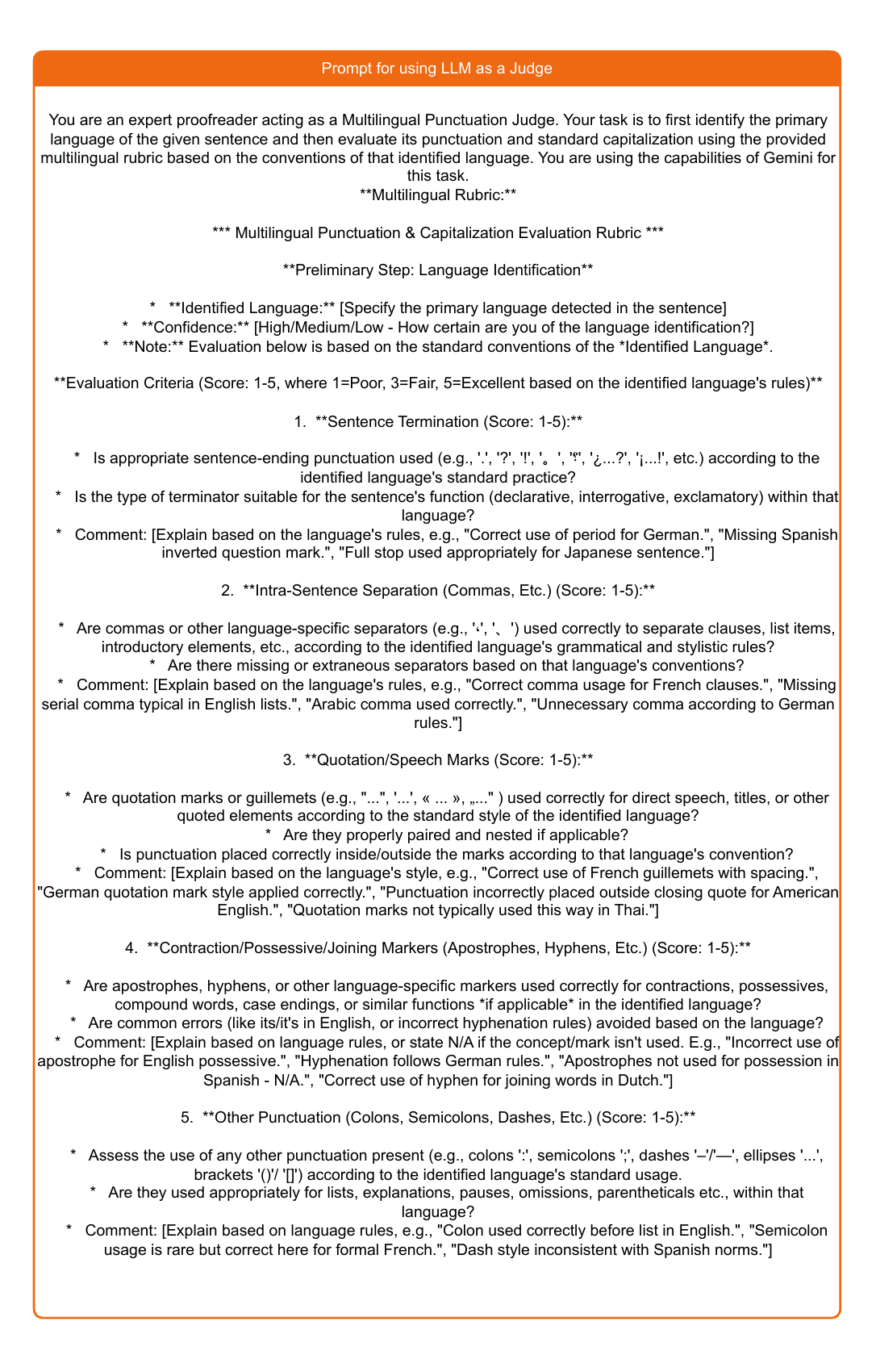}

    % \caption{The LLM-as-a-Judge prompt, outlining the comprehensive rubric used for evaluating punctuation and capitalization. Criteria include confidence, sentence termination, intra-sentence separators, quotation marks, other punctuation types (colons, semicolons, dashes), capitalization, and an overall quality assessment, along with instructions for JSON output.}
    \label{fig:llm_as_a_judge}    
\end{figure*}

\begin{figure*}
    \includegraphics[width=1.0\linewidth]{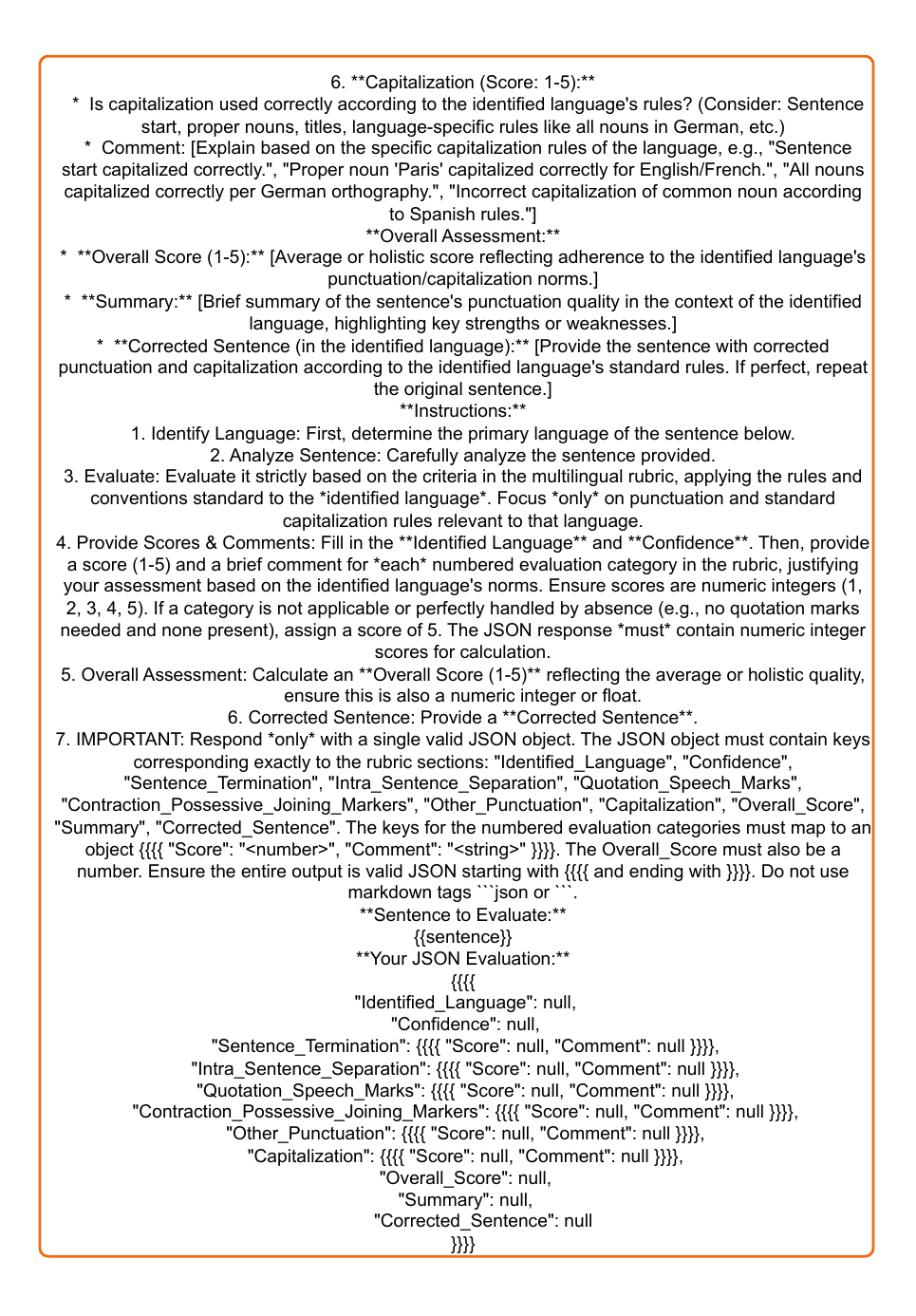}

    \caption{The LLM-as-a-Judge prompt, outlining the comprehensive rubric used for evaluating punctuation and capitalization. Criteria include confidence, sentence termination, intra-sentence separators, quotation marks, other punctuation types (colons, semicolons, dashes), capitalization, and an overall quality assessment, along with instructions for JSON output.}
    \label{fig:llm_as_a_judge}    
\end{figure*}A

%% file: main.bbl
\begin{thebibliography}{18}
\providecommand{\natexlab}[1]{#1}

\bibitem[{BehnamGhader et~al.(2024)BehnamGhader, Adlakha, Mosbach, Bahdanau, Chapados, and Reddy}]{llm2vec}
Parishad BehnamGhader, Vaibhav Adlakha, Marius Mosbach, Dzmitry Bahdanau, Nicolas Chapados, and Siva Reddy. 2024.
\newblock \href {https://openreview.net/forum?id=IW1PR7vEBf} {{LLM2V}ec: Large language models are secretly powerful text encoders}.
\newblock In \emph{First Conference on Language Modeling}.

\bibitem[{Ben~Allal et~al.(2024)Ben~Allal, Lozhkov, Penedo, Wolf, and von Werra}]{benallal2024cosmopedia}
Loubna Ben~Allal, Anton Lozhkov, Guilherme Penedo, Thomas Wolf, and Leandro von Werra. 2024.
\newblock \href {https://huggingface.co/datasets/HuggingFaceTB/cosmopedia} {Cosmopedia}.

\bibitem[{Bhogale et~al.(2025)Bhogale, Mehendale, Javed, Anuragi, Joshi, Sundaresan, Ananthanarayanan, Dey, G, Srinivasan, Raman, Kumar, and Khapra}]{10888018}
Kaushal~Santosh Bhogale, Deovrat Mehendale, Tahir Javed, Devbrat Anuragi, Sakshi Joshi, Sai Sundaresan, Aparna Ananthanarayanan, Sharmistha Dey, Sathish Kumar~Reddy G, Anusha Srinivasan, Abhigyan Raman, Pratyush Kumar, and Mitesh~M. Khapra. 2025.
\newblock \href {https://doi.org/10.1109/ICASSP49660.2025.10888018} {Towards bringing parity in pretraining datasets for low-resource indian languages}.
\newblock In \emph{ICASSP 2025 - 2025 IEEE International Conference on Acoustics, Speech and Signal Processing (ICASSP)}, pages 1--5.

\bibitem[{Devlin et~al.(2019)Devlin, Chang, Lee, and Toutanova}]{devlin2019bertpretrainingdeepbidirectional}
Jacob Devlin, Ming{-}Wei Chang, Kenton Lee, and Kristina Toutanova. 2019.
\newblock \href {https://doi.org/10.18653/V1/N19-1423} {{BERT:} pre-training of deep bidirectional transformers for language understanding}.
\newblock In \emph{Proceedings of the 2019 Conference of the North American Chapter of the Association for Computational Linguistics: Human Language Technologies, {NAACL-HLT} 2019, Minneapolis, MN, USA, June 2-7, 2019, Volume 1 (Long and Short Papers)}, pages 4171--4186. Association for Computational Linguistics.

\bibitem[{Doddapaneni et~al.(2023)Doddapaneni, Aralikatte, Ramesh, Goyal, Khapra, Kunchukuttan, and Kumar}]{doddapaneni-etal-2023-towards}
Sumanth Doddapaneni, Rahul Aralikatte, Gowtham Ramesh, Shreya Goyal, Mitesh~M. Khapra, Anoop Kunchukuttan, and Pratyush Kumar. 2023.
\newblock \href {https://doi.org/10.18653/v1/2023.acl-long.693} {Towards leaving no {I}ndic language behind: Building monolingual corpora, benchmark and models for {I}ndic languages}.
\newblock In \emph{Proceedings of the 61st Annual Meeting of the Association for Computational Linguistics (Volume 1: Long Papers)}, pages 12402--12426, Toronto, Canada. Association for Computational Linguistics.

\bibitem[{Gala et~al.(2023)Gala, Chitale, Raghavan, Gumma, Doddapaneni, M, Nawale, Sujatha, Puduppully, Raghavan, Kumar, Khapra, Dabre, and Kunchukuttan}]{gala2023indictrans}
Jay Gala, Pranjal~A Chitale, A~K Raghavan, Varun Gumma, Sumanth Doddapaneni, Aswanth~Kumar M, Janki~Atul Nawale, Anupama Sujatha, Ratish Puduppully, Vivek Raghavan, Pratyush Kumar, Mitesh~M Khapra, Raj Dabre, and Anoop Kunchukuttan. 2023.
\newblock \href {https://openreview.net/forum?id=vfT4YuzAYA} {Indictrans2: Towards high-quality and accessible machine translation models for all 22 scheduled indian languages}.
\newblock \emph{Transactions on Machine Learning Research}.

\bibitem[{Grattafiori et~al.(2024)Grattafiori, Dubey, Jauhri, Pandey, Kadian, Al-Dahle, Letman, Mathur, Schelten, Vaughan, Yang, Fan, Goyal et~al.}]{grattafiori2024llama3herdmodels}
Aaron Grattafiori, Abhimanyu Dubey, Abhinav Jauhri, Abhinav Pandey, Abhishek Kadian, Ahmad Al-Dahle, Aiesha Letman, Akhil Mathur, Alan Schelten, Alex Vaughan, Amy Yang, Angela Fan, Anirudh Goyal, and 1 others. 2024.
\newblock \href {https://arxiv.org/abs/2407.21783} {The llama 3 herd of models}.
\newblock \emph{Preprint}, arXiv:2407.21783.

\bibitem[{Guhr et~al.(2021)Guhr, Schumann, Bahrmann, and Böhme}]{deepMLP}
Oliver Guhr, Anne-Kathrin Schumann, Frank Bahrmann, and Hans~Joachim Böhme. 2021.
\newblock \href {http://ceur-ws.org/Vol-2957/sepp_paper4.pdf} {Fullstop: Multilingual deep models for punctuation prediction}.

\bibitem[{Gupta et~al.(2022)Gupta, Chhimwal, Dhuriya, Gaur, Shah, Chadha, and Raghavan}]{gupta2022indicpunctautomaticpunctuationrestoration}
Anirudh Gupta, Neeraj Chhimwal, Ankur Dhuriya, Rishabh Gaur, Priyanshi Shah, Harveen~Singh Chadha, and Vivek Raghavan. 2022.
\newblock \href {https://arxiv.org/abs/2203.16825} {indic-punct: An automatic punctuation restoration and inverse text normalization framework for indic languages}.
\newblock \emph{Preprint}, arXiv:2203.16825.

\bibitem[{Javed et~al.(2024)Javed, Nawale, George, Joshi, Bhogale, Mehendale, Sethi, Ananthanarayanan, Faquih, Palit, Ravishankar, Sukumaran, Panchagnula, Murali, Gandhi, R, M, Vaijayanthi, Karunganni, Kumar, and Khapra}]{javed2024indicvoicesbuildinginclusivemultilingual}
Tahir Javed, Janki~Atul Nawale, Eldho~Ittan George, Sakshi Joshi, Kaushal~Santosh Bhogale, Deovrat Mehendale, Ishvinder~Virender Sethi, Aparna Ananthanarayanan, Hafsah Faquih, Pratiti Palit, Sneha Ravishankar, Saranya Sukumaran, Tripura Panchagnula, Sunjay Murali, Kunal~Sharad Gandhi, Ambujavalli R, Manickam~K M, C~Venkata Vaijayanthi, Krishnan Srinivasa~Raghavan Karunganni, and 2 others. 2024.
\newblock \href {https://arxiv.org/abs/2403.01926} {Indicvoices: Towards building an inclusive multilingual speech dataset for indian languages}.
\newblock \emph{Preprint}, arXiv:2403.01926.

\bibitem[{Kakwani et~al.(2020)Kakwani, Kunchukuttan, Golla, N.C., Bhattacharyya, Khapra, and Kumar}]{kakwani2020indicnlpsuite}
Divyanshu Kakwani, Anoop Kunchukuttan, Satish Golla, Gokul N.C., Avik Bhattacharyya, Mitesh~M. Khapra, and Pratyush Kumar. 2020.
\newblock {IndicNLPSuite: Monolingual Corpora, Evaluation Benchmarks and Pre-trained Multilingual Language Models for Indian Languages}.
\newblock In \emph{Findings of EMNLP}.

\bibitem[{Khan et~al.(2024)Khan, Mehta, Sankar, Kumaravelan, Doddapaneni, B, G, Jain, Kunchukuttan, Kumar, Dabre, and Khapra}]{Khan_2024}
Mohammed Khan, Priyam Mehta, Ananth Sankar, Umashankar Kumaravelan, Sumanth Doddapaneni, Suriyaprasaad B, Varun G, Sparsh Jain, Anoop Kunchukuttan, Pratyush Kumar, Raj Dabre, and Mitesh Khapra. 2024.
\newblock \href {https://doi.org/10.18653/v1/2024.acl-long.843} {Indicllmsuite: A blueprint for creating pre-training and fine-tuning datasets for indian languages}.
\newblock In \emph{Proceedings of the 62nd Annual Meeting of the Association for Computational Linguistics (Volume 1: Long Papers)}, page 15831–15879. Association for Computational Linguistics.

\bibitem[{Loshchilov and Hutter(2017)}]{loshchilov2017decoupled}
Ilya Loshchilov and Frank Hutter. 2017.
\newblock Decoupled weight decay regularization.
\newblock \emph{arXiv preprint arXiv:1711.05101}.

\bibitem[{Penedo et~al.(2024)Penedo, Kydlíček, Sabolčec, Messmer, Foroutan, Jaggi, von Werra, and Wolf}]{penedo2024fineweb-2}
Guilherme Penedo, Hynek Kydlíček, Vinko Sabolčec, Bettina Messmer, Negar Foroutan, Martin Jaggi, Leandro von Werra, and Thomas Wolf. 2024.
\newblock \href {https://doi.org/10.57967/hf/3744} {Fineweb2: A sparkling update with 1000s of languages}.

\bibitem[{Sankar et~al.(2025)Sankar, Jain, Narasimhan, Choudhary, Suman, Khan, Kunchukuttan, Khapra, and Dabre}]{sankar2025towards}
Ashwin Sankar, Sparsh Jain, Nikhil Narasimhan, Devilal Choudhary, Dhairya Suman, Mohammed Safi Ur~Rahman Khan, Anoop Kunchukuttan, Mitesh~M Khapra, and Raj Dabre. 2025.
\newblock \href {https://openreview.net/forum?id=QYAb7tbPKZ} {Towards building large scale datasets and state-of-the-art automatic speech translation systems for 13 {I}ndian languages}.
\newblock In \emph{The 63rd Annual Meeting of the Association for Computational Linguistics}.

\bibitem[{Team et~al.(2025)Team, Kamath, Ferret, Pathak, Vieillard, Merhej, Perrin, Matejovicova, Ramé, Rivière, Rouillard, Mesnard, Cideron, bastien Grill, Ramos, Yvinec, Casbon, Pot, Penchev, Liu, Visin, Kenealy, Beyer, Zhai, Tsitsulin, Busa-Fekete, Feng, Sachdeva, Coleman, Gao, Mustafa, Barr, Parisotto, Tian, Eyal, Cherry, Peter, Sinopalnikov, Bhupatiraju, Agarwal, Kazemi, Malkin, Kumar, Vilar, Brusilovsky, Luo, Steiner, Friesen, Sharma, Sharma, Gilady, Goedeckemeyer, Saade, Feng, Kolesnikov, Bendebury, Abdagic, Vadi, György, Pinto, Das, Bapna, Miech, Yang, Paterson, Shenoy, Chakrabarti, Piot, Wu, Shahriari, Petrini, Chen, Lan, Choquette-Choo, Carey, Brick, Deutsch, Eisenbud, Cattle, Cheng, Paparas, Sreepathihalli, Reid, Tran, Zelle, Noland, Huizenga, Kharitonov, Liu, Amirkhanyan, Cameron, Hashemi, Klimczak-Plucińska, Singh, Mehta, Lehri, Hazimeh, Ballantyne, Szpektor, Nardini, Pouget-Abadie, Chan, Stanton, Wieting, Lai, Orbay, Fernandez, Newlan, yeong Ji, Singh, Black, Yu, Hui, Vodrahalli, Greff, Qiu,
  Valentine, Coelho, Ritter, Hoffman, Watson, Chaturvedi, Moynihan, Ma, Babar, Noy, Byrd, Roy, Momchev, Chauhan, Sachdeva, Bunyan, Botarda, Caron, Rubenstein, Culliton, Schmid, Sessa, Xu, Stanczyk, Tafti, Shivanna, Wu, Pan, Rokni, Willoughby, Vallu, Mullins, Jerome, Smoot, Girgin, Iqbal, Reddy, Sheth, Põder, Bhatnagar, Panyam, Eiger, Zhang, Liu, Yacovone, Liechty, Kalra, Evci, Misra, Roseberry, Feinberg, Kolesnikov, Han, Kwon, Chen, Chow, Zhu, Wei, Egyed, Cotruta, Giang, Kirk, Rao, Black, Babar, Lo, Moreira, Martins, Sanseviero, Gonzalez, Gleicher, Warkentin, Mirrokni, Senter, Collins, Barral, Ghahramani, Hadsell, Matias, Sculley, Petrov, Fiedel, Shazeer, Vinyals, Dean, Hassabis, Kavukcuoglu, Farabet, Buchatskaya, Alayrac, Anil, Dmitry, Lepikhin, Borgeaud, Bachem, Joulin, Andreev, Hardin, Dadashi, and Hussenot}]{gemmateam2025gemma3technicalreport}
Gemma Team, Aishwarya Kamath, Johan Ferret, Shreya Pathak, Nino Vieillard, Ramona Merhej, Sarah Perrin, Tatiana Matejovicova, Alexandre Ramé, Morgane Rivière, Louis Rouillard, Thomas Mesnard, Geoffrey Cideron, Jean bastien Grill, Sabela Ramos, Edouard Yvinec, Michelle Casbon, Etienne Pot, Ivo Penchev, and 197 others. 2025.
\newblock \href {https://arxiv.org/abs/2503.19786} {Gemma 3 technical report}.
\newblock \emph{Preprint}, arXiv:2503.19786.

\bibitem[{Tripathy and Samal(2022)}]{tripathy-samal-2022-punctuation}
Subhashree Tripathy and Ashis Samal. 2022.
\newblock \href {https://doi.org/10.18653/v1/2022.umios-1.9} {Punctuation and case restoration in code mixed {I}ndian languages}.
\newblock In \emph{Proceedings of the Workshop on Unimodal and Multimodal Induction of Linguistic Structures (UM-IoS)}, pages 82--86, Abu Dhabi, United Arab Emirates (Hybrid). Association for Computational Linguistics.

\bibitem[{Vandeghinste et~al.(2018)Vandeghinste, Verwimp, Pelemans, and Wambacq}]{vandeghinste-etal-2018-comparison}
Vincent Vandeghinste, Lyan Verwimp, Joris Pelemans, and Patrick Wambacq. 2018.
\newblock \href {https://aclanthology.org/2018.eamt-main.27/} {A comparison of different punctuation prediction approaches in a translation context}.
\newblock In \emph{Proceedings of the 21st Annual Conference of the European Association for Machine Translation}, pages 289--298, Alicante, Spain.

\end{thebibliography}
